\documentclass{llncs}
\usepackage{amsfonts}
\usepackage{amsmath} 
\usepackage{pifont}
\usepackage{graphicx}
\usepackage{subfigure}
\usepackage{algorithmic}
\usepackage{algorithm}
\usepackage{color}
\definecolor{articlecolor}{rgb}{1,0.5,0}
\definecolor{Lcolor}{rgb}{1,0,0.8}
\usepackage[ansinew]{inputenc}
\usepackage{array}
\usepackage{hhline}
\usepackage{fancybox}
\usepackage{subfigure}
\usepackage{wrapfig}
\usepackage[T1]{fontenc}
\usepackage{multirow}
\usepackage{color}
\usepackage{textcomp}
\usepackage{palatino} 
\usepackage{graphicx}
\usepackage{epsfig}
\usepackage{tikz}
\usepackage{soul}
\usepackage{verbatim}
\usetikzlibrary{arrows,shapes} 
\usetikzlibrary{arrows,%
                petri,%
                topaths}%
\usepackage{tkz-berge}
\usepackage{pgfplots}
\usepackage{listings}

\newcommand{\choco}{$\mathtt{Choco}$}
\newcommand{\ibex}{$\mathtt{Ibex}$}
\newcommand{\gecode}{$\mathtt{Gecode}$}

\author{Jean-Guillaume Fages \and Gilles Chabert \and Charles Prud'Homme}  
\institute{
	 TASC - Ecole des Mines de Nantes, \\
	 LINA UMR CNRS 6241,\\ 
          FR-44307 Nantes Cedex 3, France,\\
         \email{\{Jean-Guillaume.Fages,Gilles.Chabert, Charles.Prudhomme\}@mines-nantes.fr}
}                       
\title{Combining finite and continuous solvers}
\subtitle{Towards a simpler solver maintenance}
\begin{document}
\maketitle

\begin{abstract}
Combining efficiency with reliability within CP systems is one of the main concerns of CP developers.
This paper presents a simple and efficient way to connect \choco{} and \ibex{}, two CP solvers respectively specialised on finite and continuous domains. This enables to take advantage of the most recent advances of the continuous community within \choco{} while saving development and maintenance resources, hence ensuring a better software quality.
\end{abstract}

\section{Introduction}

The Constraint Programming (CP) community is witnessing the emergence of numerous new solvers, most of them coming up with new features.
In this competitive context, integrating latest advances and ensuring software quality is challenging.
From a more general point of view, spending effort on developing something already well handled by other libraries can be argued to be a waste of resource.
\choco{} \cite{Choco} and \ibex{} \cite{Ibex} are two such solvers, respectively specialised on Finite Domains (FD) and Continuous Domains (CD). While they already have some history, they have recently been completely re-engineered to brand new improved versions. 

This paper presents a bridge which has been made so that \choco{} can use \ibex{} as a global constraint. 
The interval arithmetics provided by \ibex{} greatly enhances modeling possibilities of \choco{}.
It enables to express naturally the wide family of \textit{statistical constraints} \cite{Pesant:Spread,Schaus:Deviation}, 
but also non-linear physics constraints as well as many continuous objective functions.
This bridge enables to take advantage of the most recent advances of the CD community within \choco{} for free.
It saves development and maintenance resources, and contributes to the software quality.
In this way, \choco{} and \ibex{} developers can focus on what they do best, being respectively FD and CD reasonings, and users have access to the whole.

\section{Solver overviews}

\subsection{Choco 3.0}

\choco{} is a java library for constraint satisfaction problems and constraint optimisation problems. 
This solver already has a long history and has been fully re-engineered this year, to a 3.0 version~\cite{choco32013,Choco}. It roughly contains $60,000$ lines of code.

The \choco{} library contains numerous variables, constraints and search procedures, to provide wide modeling perspectives.
Most common variables are integer variables (including binary variables and views \cite{Schulte:Views}) but the distribution also includes set variables, graph variables and real variables. 
The constraint library provided by \choco{} contains many global constraints, which gives a very expressive modeling language.
The search process can also be greatly improved by various built-in search strategies (such as DomWDeg, ABS, IBS, first-fail, etc.)
and some optimisation procedures (LNS, fast restart, etc.).
Moreover, \choco{} natively supports explained constraints.
Last, several useful extra features, such as a FlatZinc (the target language of MiniZinc~\cite{minizinc}) parser and some viewing tools, are provided as well.

\choco{} is used by the academy for teaching and research and by the industry to solve real-world problems, such as program verification, data center management, timetabling, scheduling and routing.

\subsection{Ibex 2.0}

\ibex{} (Interval-Based EXplorer) is also a library for constraint satisfaction and optimization, but written in C++ and dedicated
to continuous domains. This solver has been fully re-engineered to a 2.0 version this year~\cite{Ibex}. \ibex{} consists of roughly $40,000$ lines of code.

From the perspective of solver cooperation, two features of \ibex{} are of interest: the modeling language and the {\it contractors}.

Compared to \choco{}, the modeling language is much simpler in the sense that constraints are either numerical equations or
inequalities. However, the mathematical expression involved in a constraint can be of arbitrary complexity. The expression is obtained by composition of
standard mathematical operators such as $+$, $\times$, $\sqrt{}$, $\sin$, etc. (see \S\ref{sec:interface}).
The modeling language also allows vector and matrix operations; it shares some similarities with Matlab on purpose.

A contractor \cite{chabert-jaulin-AI-2009} is the equivalent of a propagator in finite domain except that it is considered as a pure
function: it takes a Cartesian product of domains as input and returns a subset of it. 
\ibex{} contains a variety of built-in contractors for acheiving different level of bound consistency with respect to a set of
numerical constraints such as HC4, Shaving, ACID, X-newton, q-intersection, etc.

Finally, \ibex{} also comes with a default black-box solver and global optimizer for immediate usage.
It is mainly used so far in academic labs for teaching and research. Its main application field is 
global optimization and robotics.

\section{Embedding Ibex into a Choco constraint}\label{chocobex}

\subsection{Motivation}

It is worth noticing that combining FD with CD in a CP solver is not new.
Since its early beginning, the \choco{} solver has supported real variables, hence it has always been able to solve hybrid discrete continuous problems. 
However, these older versions included their own interval arithmetics implementation.
Another example is the \gecode{} $4.0.0$ solver \cite{Gecode}, which has recently added floating variables to its distribution, by following the same approach. 

Interestingly, it appeared that most of \choco{} users and contributors were concerned by FD problems.
Thus, for historical reasons, the \choco{} module over reals has not evolved much within the last years.
In the meanwhile, people working on continuous problems have proposed new solvers, such as \ibex{},
able to handle efficiently continuous non-linear equation systems. 
As a counterpart, such solvers are not competitive on problems involving finite domains, if ever they can handle them.

If no theoretical pitfall stands in the way of implementing state-of-the-art CD techniques in \choco{}, 
this would require significant resources and ensuring its maintenance over time is presumably even more expensive. 
Moreover, it would require \choco{} developers to have a high level of expertise on both FD and CD.
A symmetric reasoning holds if one would like to implement advanced FD features within \ibex{}.
Thus, instead of reimplementing the wheel, it has been decided to make a bridge between \choco{} and \ibex{}.
This provides a very good trade-off between solver features and implementation effort.

The choice of using \ibex{} within \choco{}, instead of the opposite, is based on practical reasons.
First of all, the functional architecture of contractors in \ibex{} enables to call them from another program very easily.
Second, \choco{} has more variable types, hence using an opposite design would require a heavier interface.
In particular, \ibex{} would have to implement finite domains.
Third, \choco{} offers richer resolution options (black-box search procedures, LNS, explanations...) than \ibex{} so it is better to give the control of the search to \choco{}.
Last, calling Java from C++ is more cumbersome since a virtual machine has to be loaded prior to function calls.

\subsection{A simple but yet expressive interface}\label{sec:interface}


The bridge linking \choco{} and \ibex{} is organised in a master-slave architecture where \choco{} integrates \ibex{} within a global constraint.
This constraint, referred to as \texttt{RealConstraint}, has no particular semantics but is used as a shell to encapsulate continuous propagators.
Each equation system of the model is associated with one generic propagator, \texttt{RealPropagator}, in \choco{}  and one contractor in \ibex{}. 
Continuous expressions can embed integer variables by using views.
\choco{} drives the propagation algorithm: on domain modifications, targeted propagators are scheduled for a future execution. Any call to the propagation algorithm of a \texttt{RealPropagator} is then automatically delegated to \ibex{} contractors; the resulting domain modifications, if any, are recovered and transmitted back to \choco{}.
\ibex{} contractors are called through the Java Native Interface (JNI) which enables a Java program to call functions of a C++ library. 
Comments apart, this native interface only includes $40$ lines of code, whence the easy maintenance. An overview of the \choco{}-\ibex{} framework is given in Figure \ref{fig:bridge}. 

\tikzstyle{vertex}=[circle,draw=black]
\begin{figure}[H]
\centering
\begin{tikzpicture}[auto,scale=0.8,y=0.85cm]
  \node (RV) at (0,0) [anchor = south, draw,thick,minimum width=2cm,minimum height=1cm,align=center,text width=2cm] {\texttt{Real Variable}};
  \node (IV) at (0,2.5) [anchor = south, draw,thick,minimum width=2cm,minimum height=1cm,align=center,text width=2cm] {\texttt{Integer Variable}};
  \node (RP) at (6,0) [anchor = south, draw,thick,minimum width=2cm,minimum height=1cm,align=center,text width=2cm] {\texttt{Real Propagator}};
  \node[right] (RC) at (4,4) {\texttt{Real Constraint}};
  \draw[thick] (4,-0.5) -- (4,4.5) -- (8,4.5) -- (8,-0.5) -- (4,-0.5);
  \node[right] (CHOCO) at (-1,4.5) {CHOCO};
  \draw[thick] (-2,-1) -- (-2,5) -- (8.5,5) -- (8.5,-1) -- (-2,-1);
  \node[anchor = south, draw,thick,minimum width=2cm,minimum height=1cm,align=center,text width=2cm]  (C) at (6,-4) {\texttt{Contractor}};
  \node[right] (IBEX) at (-1,-2.5) {IBEX};
  \draw[thick] (-2,-4.5) -- (-2,-2) -- (8.5,-2) -- (8.5,-4.5) -- (-2,-4.5);
  \path[<->, thick] (RP) edge node[above] {\texttt{domains}} (RV);
  \path[<->, thick] (RV) edge node[right] {\texttt{views}} (IV);
  \path[<->, thick] (RP) edge node[right] {\texttt{JNI}} (C);
\end{tikzpicture}
\caption{Scheme of the \choco{}-\ibex{} bridge. \label{fig:bridge}}
\end{figure}

Listing \ref{propagator} provides the filtering algorithm of \texttt{RealPropagator}.
First, the propagator copies variable domain bounds in an array (l. $5-10$).
Second, it calls the \texttt{contract} method of the \ibex{} JNI class (see Listing \ref{contract}), with this array and the contractor identifier as input (l. $11-12$). 
This method updates the array of bounds in argument (for a further filtering) and returns an entailment statement. 
Third, it incorporates these changes into variable domains and, possibly, fails or becomes silent (l. $13-28$).
As any constraint of \choco{}, a \texttt{RealConstraint} can be reified.

Regarding the management of object creations and Java/C++ communication, this architecture does not bring any significant overhead.
When the first \texttt{RealConstraint} is created, the \ibex{} library is loaded once and for all by the system.
Each \ibex{} contractor is created once during the model creation, and its reference is kept in memory.
Calling an \ibex{} contractor from a \choco{} propagator has no particular overhead but the translation of the Java primitive double array which represents variable bounds to a native double array.
This takes a linear time over the number of variables that are involved, which is presumably less or equal to the contractor time complexity.

\begin{figure}[H]
\center
\lstset{
language=Java,
basicstyle=\scriptsize,
upquote=true,
aboveskip={1.5\baselineskip},
columns=fullflexible,
showstringspaces=false,
extendedchars=true,
breaklines=false,
showtabs=false,
showspaces=false,
showstringspaces=false,
identifierstyle=\ttfamily,
keywordstyle=\color[rgb]{0,0,1},
commentstyle=\color[rgb]{0.133,0.545,0.133},
stringstyle=\color[rgb]{0.627,0.126,0.941},
numbers = left,
caption = \ibex{}-based domain reduction of \texttt{RealPropagator},
label = propagator
}

\begin{lstlisting}
protected RealVar[] vars;
protected final int contractorIdx;
public void propagate(int event_mask) throws ContradictionException {
        // make variable domain bounds input array
        double domains[] = new double[2 * vars.length];
        for (int i = 0; i < vars.length; i++) {
            domains[2 * i] = vars[i].getLB();
            domains[2 * i + 1] = vars[i].getUB();
        }
        // call Ibex (note that it overwrites the input array "domains")
        int result = ibex.contract(contractorIdx, domains);
        switch (result) {
            case Ibex.FAIL: // trigger a failure
                contradiction(null, "Ibex failed");
            case Ibex.CONTRACT: // filter domains 
                for (int i = 0; i < vars.length; i++) {
                    vars[i].updateBounds(domains[2 * i], domains[2 * i + 1], aCause);
                }
                break;
            case Ibex.ENTAILED: // filter domains and become silent
                for (int i = 0; i < vars.length; i++) {
                    vars[i].updateBounds(domains[2 * i], domains[2 * i + 1], aCause);
                }
                setPassive();
                break;
            default: // do nothing
        }
}
\end{lstlisting}
\end{figure}

\begin{figure}
\center
\lstset{
language=Java,
basicstyle=\scriptsize,
upquote=true,
aboveskip={1.5\baselineskip},
columns=fullflexible,
showstringspaces=false,
extendedchars=true,
breaklines=false,
showtabs=false,
showspaces=false,
showstringspaces=false,
identifierstyle=\ttfamily,
keywordstyle=\color[rgb]{0,0,1},
commentstyle=\color[rgb]{0.133,0.545,0.133},
stringstyle=\color[rgb]{0.627,0.126,0.941},
caption = The \texttt{contract} \ibex{} function,
label = contract
}

\begin{lstlisting}
/**
 * Call the contractor cont_index associated to a continuous (in)equation system 
 * seen as a function of the form c(x_1,...,x_n), where x_1...x_n are n real variables
 * 
 * @param cont_index      - Number of the contractor (in the order of creation)
 * @param bounds - The bounds of domains under the following form:
 *                 (x1-,x1+,x2-,x2+,...,xn-,xn+), where xi- (resp. xi+) is the
 *                 lower (resp. upper) bound of the domain of x_i.
 *              
 * @return       The status of contraction or fail/entailment test. 
 *  - FAIL:  No tuple satisfies c. 
 *  - ENTAILED: The bounds of x may have been contracted. All remaining tuples satisfy c.  
 *  - CONTRACT: At least one bound of x has been reduced by more than 1%.
 *  - NOTHING: No bound has been reduced and nothing could be proven.
 */
public native int contract(int cont_index, double bounds[]);
	
\end{lstlisting}
\end{figure}

\newpage{} 
The expression of the continuous constraint (equation or inequality) is encoded in a simple {\tt String}. 
To simplify the interpretation of this {\tt String} by \ibex{}, 
variables are represented by their indices, surrounded by braces. For instance, the constraint {\tt "(\{0\}+\{1\}+\{2\})/3=\{3\}"}  
means that the fourth variable is the average of the three first ones.

This framework handles any equation system involving the following operators:
\begin{verbatim}
+, -, *, /, =, <, >, <=, >=,
sign, min, max, abs, sqr, sqrt, exp, log, pow, 
cos, sin, tan, acos, asin, atan,
cosh, sinh, tanh, acosh, asinh, atanh, atan2
\end{verbatim}
This provides wide modeling perspectives. In particular, the family of \textit{statistical} constraints, such as \texttt{Spread} \cite{Pesant:Spread} and \texttt{Deviation} \cite{Schaus:Deviation}, 
can be expressed naturally and extended by using neither monolithic ad hoc algorithms nor reformulations.  Of course, in continuous domains, equations and inequalities are ubiquitous. 

Besides being both simple and expressive, the use of {\tt Strings} enables to get very concise models.
As a counterpart, it has no safeguard against user mistakes in the declaration of continuous constraints. 
Hence building a framework which generates those {\tt Strings} may be a good perspective to make the use of this bridge safer.


\section{Practical example: using CD to express balancing}\label{example}

This section introduces the Santa Claus problem as a simple illustration of this framework.
Given a set of kids and a set of gifts, the Santa Claus problem consists of giving a gift to each child.
The average deviation of gift values must be minimised so that the gift distribution is fair.

The \choco{} model associated with this problem is given in Listing \ref{model}.
It involves integer assignment decision variables as well as real variables related to the average and the average deviation of gift prices.
In particular, the objective variable is real, hence the hybrid nature of the problem. 
On the one hand, the \texttt{AllDifferent} constraint is typically not implemented in \ibex{}, as differences have no much meaning over reals. 
On the other hand, the average and the average deviation constraints are straightforward to formulate as general \ibex{} arithmetic expressions.
Thus, we take the best from each solver.
The possibility to have real views of integer variables enables to consider integer variables within continuous systems.
Hence, even on purely integer problems, this framework makes available a wide family of constraints, for free.

\begin{figure}[H]
\lstset{
language=Java,
basicstyle=\scriptsize,
upquote=true,
aboveskip={1.5\baselineskip},
columns=fullflexible,
showstringspaces=false,
extendedchars=true,
breaklines=false,
showtabs=false,
showspaces=false,
showstringspaces=false,
identifierstyle=\ttfamily,
keywordstyle=\color[rgb]{0,0,1},
commentstyle=\color[rgb]{0.133,0.545,0.133},
stringstyle=\color[rgb]{0.627,0.126,0.941},
caption = Santa Claus \choco{} model,
label = model
}

\begin{lstlisting}
// input data
int n_kids = 3;
int n_gifts = 5;
int[] gift_price = new int[]{11, 24, 5, 23, 17};
int min_price = 5;
int max_price = 24;

// solver
Solver solver = new Solver("Santa Claus");

// FD variables
// VF is the factory for variables' declaration
IntVar[] kid_gift = VF.enumeratedArray("g2k", n_kids, 0, n_gifts, solver);
IntVar[] kid_price = VF.boundedArray("p2k", n_kids, min_price, max_price, solver);
IntVar total_cost = VF.bounded("total cost", min_price*n_kids, max_price * n_kids, solver);

// CD variable
double precision = 1.e-4;
RealVar average = VF.real("average", min_price, max_price, precision, solver);
RealVar average_deviation = VF.real("average_deviation", 0, max_price, precision, solver);

// continuous views of FD variables
RealVar[] realViews = VF.real(kid_price, precision);

// kids must have different gifts
// ICF is the factory for integer constraints' declaration
solver.post(ICF.alldifferent(kid_gift, "AC"));

// compute cost
for (int i = 0; i < n_kids; i++) {
    solver.post(ICF.element(kid_price[i], gift_price, kid_gift[i]));
}
solver.post(ICF.sum(kid_price, total_cost));

// compute the average and average deviation costs
RealVar[] allRV = ArrayUtils.append(realViews,new RealVar[]{average, average_deviation});
RealConstraint ave_cons = new RealConstraint(solver);
ave_cons.addFunction("({0}+{1}+{2})/3={3}", allRV);
ave_cons.addFunction("(abs({0}-{3})+abs({1}-{3})+abs({2}-{3}))/3={4}", allRV);
solver.post(ave_cons);

// set search strategy (selects smallest domains first)
solver.set(IntStrategyFactory.firstFail_InDomainMin(kid_gift));

// find optimal solution (the gift distribution should be fair)
solver.findOptimalSolution(ResolutionPolicy.MINIMIZE, average_deviation);
\end{lstlisting} 
%
\lstset{
language=Java,
basicstyle=\scriptsize,
upquote=true,
aboveskip={1.5\baselineskip},
columns=fullflexible,
showstringspaces=false,
extendedchars=true,
breaklines=false,
showtabs=false,
showspaces=false,
showstringspaces=false,
identifierstyle=\ttfamily,
keywordstyle=\color[rgb]{0,0,1},
commentstyle=\color[rgb]{0.133,0.545,0.133},
stringstyle=\color[rgb]{0.627,0.126,0.941},
caption = Output,
label = output
}

The output stream (Listing \ref{output}) then provides the following solution:
\begin{lstlisting}
********* Optimal solution
Kids #0 has received the gift #4 at a cost of 17 euros
Kids #1 has received the gift #3 at a cost of 23 euros
Kids #2 has received the gift #1 at a cost of 24 euros
Total cost: 64 euros
Average: 21.333333333333332 euros per kid
Average deviation: 2.8888888888888866 
\end{lstlisting} 
\end{figure}

\section{Conclusion}\label{conclusion}

We have proposed a bridge between \choco{} and \ibex{} so that \choco{} can use \ibex{} as a global constraint.
We have shown that this framework offers wide modeling possibilities while being simple and generic.
This work enables the FD and the CD communities to benefit from the work of each other and focus and their respective field of expertise.
It enables to provide a rich and reliable solver while saving development and maintenance resources.

~\\
\textbf{Acknowledgements.}
The authors thank the anonymous referees for their work and interesting comments.


\end{document}